\documentclass[letterpaper, 10 pt, journal, twoside]{IEEEtran}

\IEEEoverridecommandlockouts                              

\usepackage{times}

\usepackage{tikz}
\usepackage{comment}
\usepackage{color}

\usepackage{mathrsfs}
\usepackage{listings}
\usepackage{balance}
\usepackage{nicefrac}               
\usepackage{setspace}
\usepackage{xspace}
\usepackage{bbm}
\usepackage{makecell}
\usepackage{subfigure}

\usepackage{graphicx}
\usepackage{amsmath}
\usepackage{amssymb}
\usepackage{dblfloatfix}
\usepackage{transparent}
\usepackage{booktabs}
\usepackage{tabulary}

\usepackage{url}
\usepackage{multirow}
\usepackage{array}
\usepackage{bbding}
\usepackage{wrapfig}
\usepackage{colortbl}
\usepackage{indentfirst} 
\usepackage{pgfplots}
\usepackage{cancel}

\usepackage{bm}
\newcommand{\myparagraph}[1]{{ \noindent \bf #1}}
\renewcommand{\myparagraph}[1]{{\vspace{.1em} \noindent \bf #1}}

\newcommand{\etal}{\textit{et al.}}

\newcommand{\ie}{\textit{i}.\textit{e}.}
\newcommand{\eg}{\textit{e}.\textit{g}.}
\newcommand{\vs}{\textit{vs}.}

\newcommand{\blue}{\textcolor{blue}}
\definecolor{mygreen2}{RGB}{0, 139, 0}
\newcommand{\green}{\textcolor{mygreen2}}

\usepackage{algorithm2e}

\definecolor{codegreen}{rgb}{0,0.5,0}
\definecolor{codeblue}{rgb}{0.25,0.5,0.5}
\definecolor{codegray}{rgb}{0.6,0.6,0.6}
\definecolor{codecomment}{rgb}{0.059,0.43,0.54} 

\usepackage{pifont}
\newcommand{\cmark}{\ding{51}}%
\usepackage{hyperref}       

\makeatletter
\renewcommand{\maketag@@@}[1]{\hbox{\m@th\normalsize\normalfont#1}}%
\makeatother

\title{3D Part Assembly Generation with \\ Instance Encoded Transformer}

\author{Rufeng Zhang$^{\dagger\star}$, Tao Kong$^{\star}$, Weihao Wang, Xuan Han and Mingyu You$^{\ddagger}$
\thanks{Manuscript received February, 24, 2022; Revised May, 23, 2022; Accepted June, 20, 2022.}
\thanks{This paper was recommended for publication by Editor Cesar Cadena Lerma upon evaluation of the Associate Editor and Reviewers' comments.}%
\thanks{Rufeng Zhang, Weihao Wang, Xuan Han and Mingyu You are with the College of Electronic and Information Engineering, Frontiers Science Center for Intelligent Autonomous Systems, Tongji University, Shanghai, China.
        {\tt\small \{cxrfzhang, wwhtju, hanxuan, myyou\}@tongji.edu.cn}. Tao Kong is with ByteDance AI Lab, Beijing, China.
        {\tt\small kongtao@bytedance.com}.}%
\thanks{
$^{\dagger}$Intern at ByteDance. $^{\star}$Equal contribution.
 $^{\ddagger}$Corresponding author.}%
\thanks{Digital Object Identifier (DOI): see top of this page.}
}

\begin{document}

\maketitle

\begin{abstract}

It is desirable to enable robots capable of automatic assembly.
Structural understanding of object parts plays a crucial role in this task yet remains relatively unexplored.
In this paper, we focus on the setting of furniture assembly from a complete set of part geometries, which is essentially a 6-DoF part pose estimation problem.
We propose a multi-layer transformer-based framework that involves geometric and relational reasoning between parts to update the part poses iteratively.
We carefully design a unique instance encoding to solve the ambiguity between geometrically-similar parts so that all parts can be distinguished.
In addition to assembling from scratch, we extend our framework to a new task called in-process part assembly.
Analogous to furniture maintenance, it requires robots to continue with unfinished products and assemble the remaining parts into appropriate positions.
Our method achieves far more than 10\% improvements over the current state-of-the-art in multiple metrics on the public PartNet dataset.
Extensive experiments and quantitative comparisons demonstrate the effectiveness of the proposed framework.


\end{abstract}


\section{Introduction}

\IEEEPARstart{A}{utomatic} assembly has excellent potential in 3D computer vision and is also a desirable functionality for many intelligent robot systems.
Yet assembling a well-connected and structurally-stable 
IKEA furniture from scratch, even for human beings, is a cumbersome and time-consuming process.
The task of 3D part assembly involves an extremely large solution space when there is no instruction manual or step-by-step video demonstration as guidance.

\begin{figure}[!t]
\centering
\includegraphics[width=0.45\textwidth]{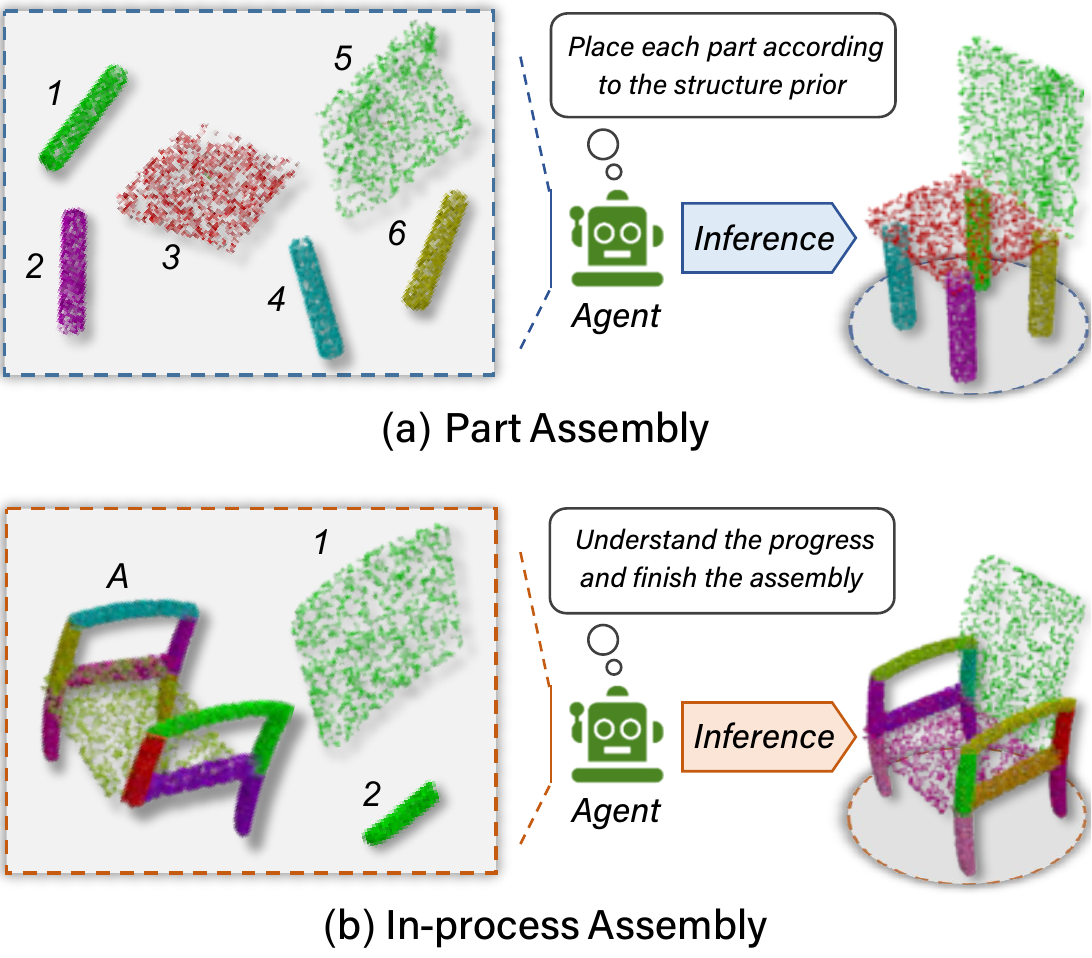}
\caption{\textbf{Illustration of two basic tasks}.
(a) is the regular task which aims to assemble objects from scratch.
(b) indicates the proposed in-process task that requires both the understanding of unfinished furniture and the relationship modeling of the  remaining parts.}
\label{fig:task_intro}
\end{figure}

Given geometry information (\ie, part point cloud) of each part, we focus on accurately estimating the 6-DoF posture 
of these parts so that they can be composed into a complete object (\eg, furniture, see Fig.~\ref{fig:task_intro}(a)). 
Compared with naive pose estimation, it requires joint prediction of all partial poses in part assembly. As part of the whole, part poses follow a series of rigid connections and relationships, such as symmetry and parallelism.
Unlike many structure-aware shape modeling works~\cite{li2020learning2, schor2019componet, wang2018global, wu2019sagnet} that specify the semantics or granularity of the input parts in advance, robots in our setting may lack 
these strong priors of the provided furniture.
In other words, the number of candidate parts is arbitrary and the semantics of each part is unknown, which puts forward higher requirements for the generalization and robustness of the robot system.

To achieve this, we present a transformer-based~\cite{vaswani2017attention} framework to predict the 6-DoF pose for each part, including rigid rotation and translation. In assembly, we employ the self-attention mechanism to capture the part relationship constraints and the part poses are gradually refined in a coarse-to-fine manner. 
Several works~\cite{li2020learning,huang2020generative,narayan2022rgl} have attempted to tackle this problem lately.
For example, Li~\etal~\cite{li2020learning} propose to assemble 3D parts guided by an additional image. 
And Huang~\etal~\cite{huang2020generative} construct a dynamically varying part graph to induce the 6-DoF poses.
These algorithms consistently advance the field, yet the performance still fails in household scenario.
They typically focus on the design of relational-reasoning network architecture while ignoring the \textit{instance difference} between parts.
Take a chair with four legs for example. On the one hand, although the legs share the same geometry shape, they differ in practical assembly pose. The geometry-identical input parts usually result in intra-class conflicts and failure in accurate prediction of differentiated poses.
On the other hand, some parts may have similar geometrical structures. The square-shaped part could be a chair seat or chair back; distinguishing the geometric subtlety of inter-class details leads to a struggling problem.

Inspired by such findings, we propose \textit{Instance Encoding}, which considers both intra-class and inter-class (dis)similarity. Correspondingly, inter-class encoding aims to
make all parts distinguishable while intra-class encoding emphasizes the relationships among those identical parts.
The encoding algorithm is designed to be independent of the order or attributes of the input parts. It allows the model to attend relative discrepancy effortlessly.
More details are presented in the later section.
At the core of our method, the encoded instance vector is attached to the self-attention module to alleviate the confusion between parts.



Besides assembling from scratch, to continue with unfinished products is also a crucial skill for robots to generalize the applicability in reality, such as furniture maintenance. We term the new task as in-process part assembly (see Fig.~\ref{fig:task_intro}(b)). Taking a piece of in-process furniture and several unassembled parts as input, the robot needs to make a smooth and collision-free assembly.
This task is particularly challenging as it requires a prior structural understanding of in-process shape and then placing the new parts in their appropriate locations. To this end, we further extend our framework using an encoder-decoder paradigm. The unfinished furniture is fed to the encoder to generate feature memory and the remaining parts query the information from the memory through the decoder. To the best of our knowledge, our algorithm is the first to assemble furniture with an in-process inventory, and the surplus part.

Following~\cite{huang2020generative, narayan2022rgl}, we demonstrate the effectiveness of our approach on the finest-grained part granularity in the large-scale PartNet~\cite{mo2019partnet} dataset. 
\textit{Instance Encoding} increases the network’s capacity to learn intricate structures, boosting the performance tremendously. Overall, our model achieves far more than 10\% improvements in both part accuracy and connectivity accuracy over SOTA model RGL-Net~\cite{narayan2022rgl}, which employs a time-consuming progressive strategy.
Extensive experiments also address the critical aspects of our scheme, including the architectural design and the advantage/limit of \textit{Instance Encoding}.
The main contributions of this work can be summarized as follows.

\begin{itemize}
\itemsep 2pt
    \item %
    We present a novel transformer-based framework involving geometric and relational reasoning between parts for 3D part assembly, 
    surpassing existing best results by a large margin.
    
    \item %
    We propose to encode instance-aware information in the task of part assembly. The coding vector takes into account both intra-class and inter-class (dis)similarity, significantly improving the accuracy of part placement and part connectivity.
    

    \item %
    We introduce a novel task termed in-process part assembly, along with the learning-based solution. Different from existing frameworks that always require the robot to assemble from scratch, the new problem emphasizes the assembly from an intermediate state. And the extensibility of our framework considerably generalizes its applicability in real life.
    
\end{itemize}

\section{Related Work}


\myparagraph{Assembly-based 3D modelling.}
Automated part assembly is a long-standing research challenge.
Previous works involving part assembly could be roughly categorized into two groups.
Several methods
~\cite{litvak2019learning, shao2020learning, zachares2021interpreting}
emphasize motion planning, actuator control, and 6-DoF grasping in robotics. Specifically, Litvak~\etal~\cite{litvak2019learning} propose a two-stage pose estimation pipeline to learn robotic assembly with depth images. Shao~\etal~\cite{shao2020learning} utilize fixtures to help the robot learn manipulation skills autonomously. To ensure the uncertainty of robust sensorimotor control, Zachares~\etal~\cite{zachares2021interpreting} formulate multi-object assembly in a hierarchical way.

The second family of solutions focuses on the problem of pose or joint estimation for part assembly, which has some similarities with approaches in the vision and graphics community. To our knowledge,~\cite{funkhouser2004modeling} is the first to construct new 3D geometric objects by assembling parts from a large household database. Several works~\cite{chaudhuri2011probabilistic,jaiswal2016assembly} use the probabilistic graph to learn semantic and stylistic relationships between components in a shape repository.
More recently, Dubrovina~\etal~\cite{dubrovina2019composite} propose to model the semantic structure-aware 3D shape for part-level shape manipulation. PAGENet~\cite{li2020learning2} introduces a part-aware network to generate semantic parts and estimate joint poses sequentially.

These methods either rely on the third-party shape repository to query parts, or assume strong priors of structure-variable (\ie, the component is not rigid and can
be arbitrarily scaled and distorted) or semantic-aware (\eg, for a particular chair, it has four legs, a backrest, a seat and two arms).
Here we tackle the problem in a more practical setting. 
No shape database is provided, no semantic information is known, no part structure could be tuned.
Instead, it takes arbitrary parts as input without any semantic knowledge, and estimates per-part poses jointly for the plausible and structurally stable shape assembly.
Several works~\cite{li2020learning, huang2020generative, narayan2022rgl} formulate a similar task.
Specifically, DGL-Net~\cite{huang2020generative} iteratively refines the part poses with a dynamic part graph. Narayan~\etal~\cite{narayan2022rgl} explore a progressive strategy via the recurrent graph learning framework.
These algorithms solve the assembly task as graph-level predictions, while we employ Transformer~\cite{vaswani2017attention} to model the structural relationships. As far as we know, we are the first to demonstrate the effectiveness of Transformer in the task of 3D part assembly.

\begin{figure*}[!t]
\centering
\includegraphics[width=0.85\textwidth]{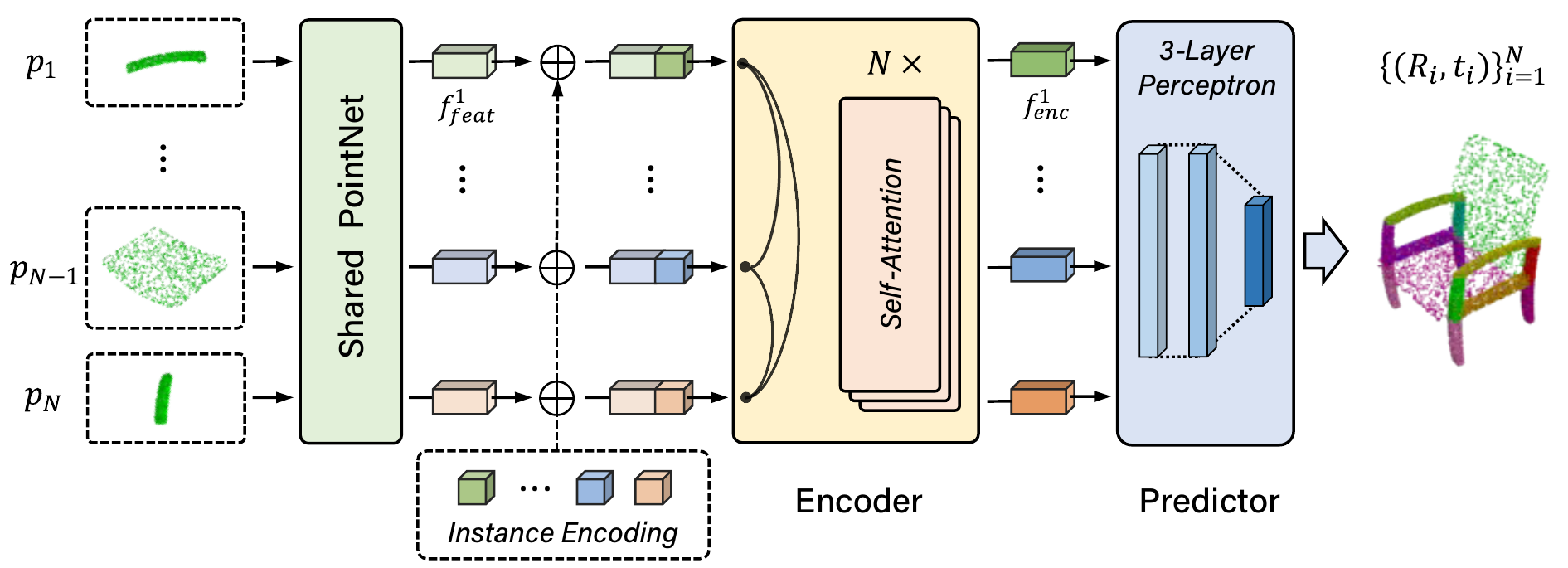}
\caption{\textbf{The overall architecture of our method}, which mainly consists of four modules: (a) shared PointNet for feature extraction, (b) transformer-encoder to reason the inter-part relationships, (c) MLP predictor for pose estimation and (d) \textit{Instance Encoding} to make input parts distinguishable. Here $\bigoplus$ denotes the concatenate operation of part feature and encoded vector.}
\label{fig:enc_arch}
\vspace{-0.48em}
\end{figure*}

\myparagraph{Structure-aware shape generation.}
Deep generative models~\cite{goodfellow2014generative,kingma2013auto} are the powerful way to shape generation tasks and have achieved tremendous success in just a few years.
For example, GRASS~\cite{li2017grass} employs a generative recursive autoencoder to encode part structure in a hierarchical way. Wang~\etal~\cite{wang2018global} present a global-to-local GAN-based paradigm that performs overall-structure generation and per-part segmentation. Moreover,~\cite{mo2020structedit} extends the learned semantically plausible variations into structure editing through a conditional variational autoencoder (VAE). Taking semantic-aware parts as input, SAGNet~\cite{wu2019sagnet} introduces a structure-aware generative model to jointly learn the part geometry as well as the pairwise relationships between parts.~\cite{gao2019sdm} decomposes the global shape structure and generates fine-grained part geometries using a two-level VAE. Recently, Mo~\etal~\cite{mo2020pt2pc} propose to generate 3D point cloud shapes with geometric variations conditioned on the part-tree hierarchy.
Most of these works directly map new part shape from the latent code, while we focus on the rigid transformation of existing parts for part assembly.

\section{Our Method}

Let $\mathcal{P} = \{p_i\}_{i=1}^{N}$ represent a set of part point clouds, where $p_{i}\in\mathbb{R}^{n_{pc}\times{3}}$ and $N$ denotes the number of parts which may vary for different 3D shapes.
Our goal is to predict a set of 6-DoF part poses $\mathcal{T} = \{(R_{i}, t_{i})\}_{i=1}^{N}$ in $SE(3)$ space, where $R_{i}\in\mathbb{R}^{4}$ and $t_{i}\in\mathbb{R}^{3}$ denotes the rigid rotation and translation for each part, respectively. We use unit quaternion to represent rotation, \ie, $\|R_{i}\|_{2} = 1$.
And the complete shape can be assembled into $S = T_{1}(p_{_1}) \cup T_{2}(p_{_2}) \cup \cdots \cup T_{N}(p_{_N})$, in which $T_i$ is the joint transformation of $(R_{i}, t_{i})$.

In this work, we present a multi-layer transformer-based framework to assemble 3D shapes in a coarse-to-fine manner. We apply the self-attention module to reason the inter-part relationships.
And \textit{Instance Encoding} is introduced to resolve ambiguity between parts, in which case some parts may share similar geometries. 
In addition, we explore a memory-query paradigm to process the in-process inventory. 
The overall pipeline is illustrated in Fig.~\ref{fig:enc_arch}.

\subsection{Instance Encoded Transformer}
Our system takes a set of 3D point clouds as input and jointly estimates the poses of parts in the space of $SE(3)$. 
A part cloud is an unordered set of $n_{pc}$ points, where each point is associated with its 3-dimensional XYZ coordinates.
We first use a shared PointNet~\cite{qi2017pointnet} to extract the  global permutation-invariant feature $f_{feat}^{i}$ for each input part $p_i$. Then these part features are sent into transformer-encoder to model the inter-part relations, denoted as $\{f_{enc}^{i}\}_{i=1}^{N}$. We apply a simple feed-forward network (FFN) to make the final pose estimation.
As for the task of in-process part assembly, the candidate part $f_{dec}^{j}$ is processed with transformer-decoder and queries the adjacent information from the part encoder-memory $\{f_{enc}^{i}\}_{i=1}^{N'}$ (here $N' < N$).

\myparagraph{Instance encoding.}
Since both the PointNet and Transformer extract features in a permutation-invariant way, it inevitably falls into the dilemma of inter-part conflict, resulting in poses failure.
For example, the geometry of the four legs of a chair is identical, and it is difficult for the network to distinguish which portion is in the upper-left corner and which one should be the lower-right leg.
Besides, some parts, such as the chair backrest and chair seat, are very similar in appearance (\eg, their geometries are both square-shaped), and the robot may make the opposite judgments.

To alleviate this confusion, we propose to encode each part with a unique instance vector, termed \textit{Instance Encoding}.
Specifically, we first cluster the input part clouds $\{p_i\}_{i=1}^{N}$ into sets of geometrically-equivalent part classes $\mathcal{C} = \{C_1, C_2, \cdots, C_k\}$, in which $C_1 = \{p_i\}_{i=1}^{N_1}$, $C_2 = \{p_i\}_{i=N_{1}+1}^{{N_1}+{N_2}}$, etc.
Following~\cite{li2020learning}, we calculate axis-aligned bounding boxes (AABB) for all the parts and measure the similarity of these 3D boxes.
The parts will be grouped together when the difference between the corresponding enclosing boxes is less than a certain threshold (\eg, $0.1$).
In the case of a chair, the four legs are clustered together as they have identical geometries. Note $\mathcal{C}$ is a disjoint complete set, \ie, $C_k \cap C_l = \varnothing$ when $k \neq l$ and $\cup_{k=1}^{K}C_{k} = \mathcal{P}$.
\textit{Instance Encoding} consists of two terms: inter-class encoding and intra-class encoding. 
In inter-class encoding, we treat each part as an individual category and encode it with a one-hot vector in the length of $N$. It ensures that each part has a unique property, making it easy to distinguish.
In intra-class encoding, geometrically-equivalent parts are attached with the same code, which enforces the connections among these highly-correlated parts.
\textit{Instance Encoding} is simple but efficient, and we demonstrate it is the key to distinguishing indiscernible parts in the task of part assembly. 
We detail the generation procedure in Algorithm~\ref{alg:instance_encoding}.
\RestyleAlgo{ruled}


\begin{algorithm}[t]
\caption{Instance Encoding}
\label{alg:instance_encoding}
\KwIn{\\
\quad 3D point clouds, $\mathcal{P} = \{p_i\}_{i=1}^{N}$; \\
\quad Geometrically-equivalent classes, $\mathcal{C} = \{C_k\}_{k=1}^{K}$.}
\KwOut{\\
\quad Instance encoding set, $\mathcal{V} = \{\}$.}
\For{$i = 1;\ i \leq N;\ i = i + 1$}{
    $v_{inter}^{i} = f_{one\_hot}(i)$\;
    $k\gets 1$\;
    \While{$k\leq K$}{
    \eIf{$p_i\in C_k$}{
    $v_{intra}^{i} = f_{one\_hot}(k)$\;
    $v_{i}\gets (v_{inter}^{i}, v_{intra}^{i})$\;
    Add $v_{i}$ to $\mathcal{V}$\;
    Break\;}{
    $k\gets k+1$\;}
}
}
\end{algorithm}

\myparagraph{Encoder.}
We employ the transformer-encoder to learn the relationships between parts. The encoder applies multiple self-attention layers that aggregate information (\eg, geometry and posture) from the entire input sequence (here is a set of parts), thus making it an optimal option for part assembly. We follow the standard formulation of transformer-encoder, and we refer the reader to Vaswani~\etal~\cite{vaswani2017attention} for details. 
The positional encoding~\cite{bello2019attention} is omitted as the input already contains the information about 3-dimensional XYZ coordinates.
Instead, \textit{Instance Encoding} is attached to the input of each attention layer to make parts distinct.

\myparagraph{Predictor.}
We carry out a 3-layer perceptron for the final prediction.
Specifically, the first two fully-connected layers have the channel
of 256, followed by ReLU activation function.
And the last linear projection layer is used to estimate the part pose $T_i \in \mathbb{R}^{7}$, including the 4-dimensional rigid rotation and 3-dimensional translation. 
We apply \texttt{tanh} operation to the translation vector, constraining the part center offset to (-1, 1).
To ensure the output of unit Quaternion prediction, we normalize the rotation vector so that $\|R_{i}\|_{2} = 1$.
The feed-forward network (FFN) works on all outputs of encoder layers and the parameters are shared.
Following~\cite{huang2020generative,narayan2022rgl}, the previous predicted poses are attached to the next step for more coherent pose evolution.

\begin{figure}[!t]
\centering
\includegraphics[width=0.48\textwidth]{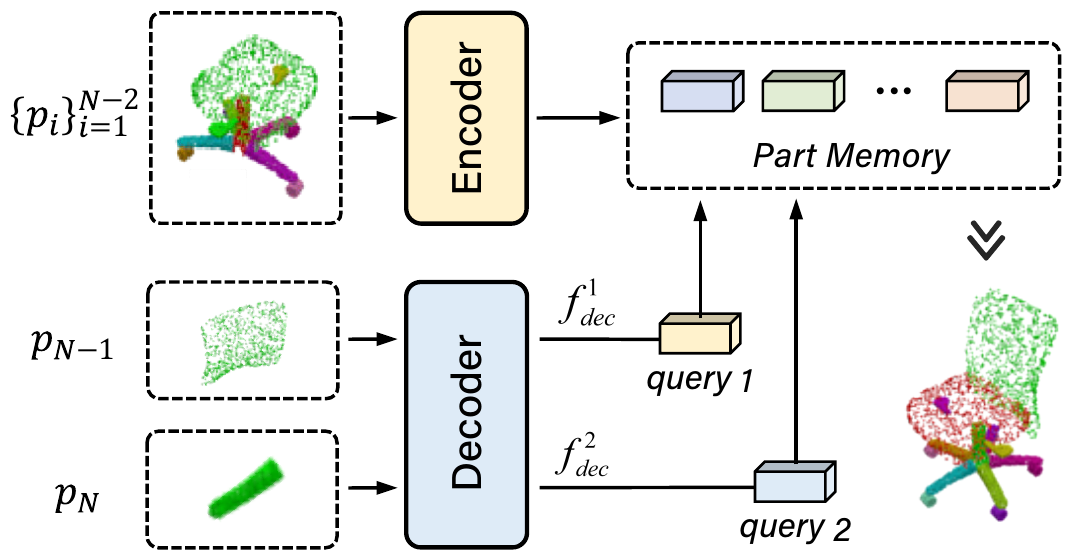}
\caption{\textbf{An overview of in-process assembly pipeline}.
It consists of two branches: encoder branch and decoder branch.
The input includes a piece of unfinished object and several parts.
The incomplete product is fed to encoder to generate part feature memory.
And the remaining parts first query the relation information from memory and then output their poses.}
\label{fig:dec_arch}
\end{figure}

\myparagraph{Decoder.}
Actually, the aforementioned modules have already been well qualified for the task of part assembly.
In this subsection, we extend the application of our framework to a new task, termed in-process part assembly. Previous works always use scattered parts and assemble them into a complete shape from scratch.
While sometimes an object is already being processed, we only need to assemble the remaining parts.
We append the transformer-decoder to tackle this problem.
The decoder takes candidate parts as input, and queries the relationship information from in-process shapes (these features have been obtained through the encoder).
The part queries are transformed into output embeddings by the decoder.
And they are then decoded into poses using the feed-forward predictor (described in the last subsection).
With self- and encoder-decoder attention over these embeddings, the model globally reasons about all parts together using pair-wise relations between them, while being able to assemble new parts to in-process shape smoothly.
The pipeline is shown in Fig.~\ref{fig:dec_arch}.

\myparagraph{PartDrop.}
In a standard paradigm of part assembly, all parts are typically input to the encoder for relation modeling.
However, there is an information vulnerability in the in-process object as it has no access to capture the holistic shape.
If we directly use the pre-trained part-assembly encoder, the model fails to support incomplete-shape memory because of the information gap between these two tasks.
Therefore, we adopt a simple solution to this problem.
The input parts are randomly dropped during training so that the model learns to capture in-process shape features.
\textit{PartDrop} largely improves the capacity of encoder in the case of in-process objects.

\subsection{Training and Losses}
In the task of part assembly, solutions are usually not unique but have multiple potential options. The locations of geometrically-equivalent parts are interchangeable, and the decorative portion could be placed in any suitable corner.
Similar to~\cite{huang2020generative,narayan2022rgl}, we employ the Min-of-N (MoN) loss~\cite{fan2017point} for uncertainty modeling, and random noise is incorporated to explore structural diversity.
We conduct Hungarian matching for all the geometrically-equivalent part groups.
Considering the overall framework as $\mathcal{F}$ and the ground-truth poses as $\mathcal{F}^{*}$, we define the MoN loss as:
\begin{equation}
\label{eq:equ_mon}
\mathcal{L}_{MoN} = \min_{
    \substack{r_{j}\sim\mathcal{N}(0,1) \\ 1 \leq j \leq n}}
    \mathcal{L}(\mathcal{F}(\mathcal{P},r_{j}), \mathcal{F^{*}}(\mathcal{P}))
\end{equation}
where $r_j$ is the IID random noise sampled from unit Gaussian distribution $\mathcal{N}(0,1)$.
Given a set of part point clouds $\mathcal{P}$, $\mathcal{F}$ makes $n$ predictions by perturbing the input
with $n$ random vectors $r_j$.
Intuitively, it ensures at least one prediction as close as the ground-truth space.
Following~\cite{huang2020generative,narayan2022rgl}, we set $n = 5$ in the experiment and supervise the model with global and part-wise losses.

We apply the Euclidean loss to measure the distance between the predicted translation $t_i$ and ground-truth translation $t_{i}^{*}$ for each part, formally, 
\begin{equation}
\label{eq:equ_trans}
\mathcal{L}_{t} = \sum_{i=1}^{N}\|t_i - t_i^*\|_{2}^{2}
\end{equation}

And Chamfer distance (CD) loss is used to supervise rigid rotation:
\begin{equation}
\label{eq:equ_rot}
\footnotesize
\mathcal{L}_{r} = \sum_{i=1}^{N}\bigg(
\sum_{x \in R_{i}(p_i)}\min_{\substack{y \in \\ R^{*}_{i}(p_i)}}\|x - y\|_{2}^{2}
+ \sum_{y \in R^{*}_{i}(p_i)}\min_{\substack{x \in \\ R_{i}(p_i)}}\|x - y\|_{2}^{2}\bigg)
\end{equation}
in which $R_{i}(p_i)$ and $R^{*}_{i}(p_i)$ represent the rotated part points $p_i$ using the estimated rotation $R_{i}$ and the ground-truth $R^{*}_{i}$, respectively. 
The loss searches the nearest neighbor in the other set for each point, enabling the tolerance to part symmetry.
Finally, we define the global CD loss to learn the holistic assembled shape $S$:
\begin{equation}
\label{eq:equ_shape}
\mathcal{L}_{s} = \sum_{x \in S}\min_{y \in S^{*}}\|x - y\|_{2}^{2}
+ \sum_{y \in S^{*}}\min_{x \in S}\|x - y\|_{2}^{2}
\end{equation}
where $S$ is the assembled shape and $S^{*}$ denotes the ground truth.
The overall loss is defined as follows:
\begin{equation}
\label{eq:equ_total}
\mathcal{L} = \lambda_{t}\mathcal{L}_{t} + \lambda_{r}\mathcal{L}_{r} + \lambda_{s}\mathcal{L}_{s}
\end{equation}

We extend the part assembly task to in-process assembly task by simply adding a decoder module to our framework. The parameters in encoder are frozen and initialized with the pre-trained part-assembly model.
We randomly sample one part in a complete shape and extract its feature using decoder. The remains are input to encoder to generate in-process object memory. 
The loss functions are similar to the ones in part assembly.

\subsection{Evaluation Metrics}

\myparagraph{Part assembly.}
Similar to~\cite{huang2020generative,narayan2022rgl}, we make multiple predictions by adding different noises to the input parts and find the most similar shape to ground truth using Minimum Matching Distance (MMD)~\cite{achlioptas2018learning}.
The quality of assembled shapes is evaluated with three metrics: \textit{shape Chamfer distance}, \textit{part accuracy} and \textit{connectivity accuracy}. \textit{Shape Chamfer distance} (SCD) can be directly obtained through Eq.~\eqref{eq:equ_shape}. And we use \textit{part accuracy} (PA) to measure the percentage of matched parts within a certain Chamfer distance threshold, formally,

{\begin{small}
\setlength{\abovedisplayskip}{-5pt}
\begin{align}
\mathcal{PA} = \dfrac{1}{N}\sum_{i=1}^{N}\bigg(\Big(
\sum_{x \in R_{i}(p_i)}\min_{y \in R^{*}_{i}(p_i)}\|x - y\|_{2}^{2}
\notag \\ + \sum_{y \in R^{*}_{i}(p_i)}\min_{x \in R_{i}(p_i)}\|x - y\|_{2}^{2}\Big) < \tau_{p} \bigg),
\label{eq:equ_pa}
\end{align}
\end{small}}

\begin{flushleft}{where we set the threshold $\tau_{p} = 0.01$.}
\vspace{-0.6em}
\end{flushleft}

Finally, \textit{connectivity accuracy} (CA) is introduced to evaluate the quality of connections between adjacent parts.
For each connected-part pair $(p_i^{*},p_j^{*})$ in the object space, we firstly find the point in part $p_i^{*}$ closest to $p_j^{*}$ and define it as contact point $c_{ij}^{*}$, so does $c_{ji}^{*}$.
The contact-point pair $(c_{ij}^{*}, c_{ji}^{*})$ is transformed into part canonical space as $(c_{ij}, c_{ji})$, correspondingly.
We formulate the connectivity accuracy as Eq.~\eqref{eq:equ_ca}:

\begin{equation}
\label{eq:equ_ca}
\setlength{\abovedisplayskip}{-5pt}
\mathcal{CA} = \dfrac{1}{\left|\mathcal{C}\right|}\sum_{\{c_{ij}, c_{ji} \in \mathcal{C}\}} \|T_{i}(c_{ij}) - T_{j}(c_{ji})\|_{2}^{2} < \tau_{c}
\end{equation}
where $\mathcal{C}$ is the set of all contact point pairs in a shape and $\tau_{c} = 0.01 $ denotes the threshold of Chamfer distance.

\myparagraph{In-process part assembly.}
We employ the same evaluation metrics for the in-process part assembly task.
We sample one part from a complete shape in turn and measure the performance of assembled part individually. The result is the average of all candidate parts in a shape.
Formally,
\begin{equation}
\label{eq:equ_wip}
\dfrac{1}{N}\sum_{i=1}^{N}\mathcal{D}\big(T_{i}(p_{i}) \cup S_{i}^{*}\big)
\end{equation}
where $S_{i}^{*}$ denotes the ground-truth incomplete shape that excludes part $p_{i}$ and $\mathcal{D}$ can be any metric described above.

\begin{figure}[!t]
\centering
\includegraphics[width=0.42\textwidth]{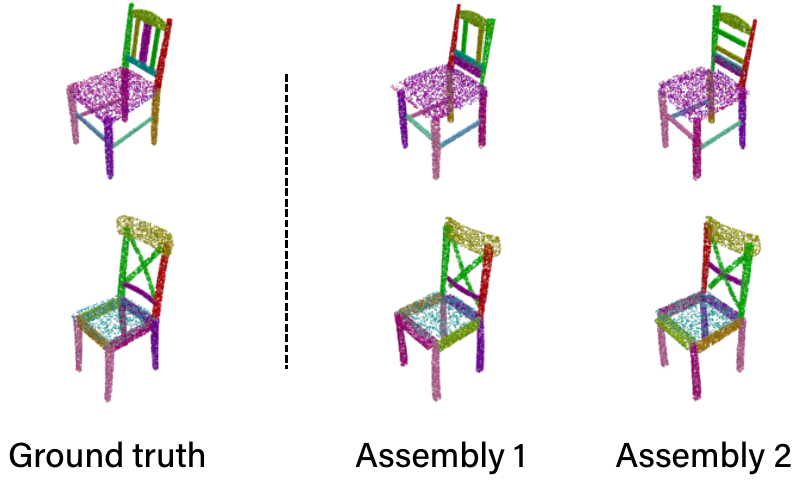}
\setlength{\abovecaptionskip}{-0.01cm} 
\caption{\textbf{Multiple plausible assembly results}. 
The assembled chairs are not exactly the same as the ground truth, yet their structure is reasonable.
}
\vspace{-0.36em}
\label{fig:multi_poses}
\end{figure}

\section{Experiments}

\subsection{Dataset}

Our experiments are conducted on PartNet~\cite{mo2019partnet} dataset using the metrics described above.
PartNet has 24 categories with fine-grained part annotations, and we select the three largest categories (Chair, Table, and Lamp) for both training and evaluation.
In total, we have 6,323 chairs, 8,218 tables, 2,207 lamps in finest-grained level, and we follow the default train/validation/test splits (70\%/10\%/20\%) in the dataset.
Furthest Point Sampling (FPS) is adopted to sample 1,000 points for each part cloud and we filter out the objects with more than 20 parts. All input parts are normalized within the canonical space using PCA.

\subsection{Implementation Details}

\myparagraph{Train details.}
We use the AdamW optimizer, initial learning rate 0.00015, weight decay 0.0001 with 1,000 epochs in all for the task of part assembly. We use a mini-batch of 64 and all models are trained with 8 GPUs.
The default number of encoder layers is 6 and intermediate supervision is applied to all the output poses of encoders to accelerate the convergence.
Following~\cite{huang2020generative}, $\lambda_{t}=1$, $\lambda_{r}=10$, $\lambda_{s}=1$.

As for in-process part assembly, we add a 6-layer decoder to learn the remaining parts.
We initialize this model with a pre-trained part-assembly model, then freeze all weights in encoder and train only the decoder head for 500 epochs.

\myparagraph{Inference details.}
Unlike training, we only use the final estimated poses to measure the quality of assembled shapes.
To make a fair comparison with the previous methods~\cite{huang2020generative, narayan2022rgl}, we generate 10 possible shapes and find the prediction most similar to ground truth using Minimum Matching Distance (MMD).

\begin{table*}[!t]	
	\centering
	\small
	\setlength{\abovecaptionskip}{0cm}
    \caption{Part assembly results on PartNet dataset.}
    \setlength{\tabcolsep}{6pt}
\centering
\begin{tabular}{c ccc | ccc | ccc}
\toprule
& \multicolumn{3}{c}{Shape Chamfer Distance $\downarrow$} & \multicolumn{3}{c}{Part Accuracy $\uparrow$} & \multicolumn{3}{c}{Connectivity Accuracy $\uparrow$} \\ 
\midrule
& Chair & Table & Lamp & Chair & Table & Lamp & Chair & Table & Lamp \\
\midrule				
B-Global~\cite{li2020learning2, schor2019componet} & 0.0146 & 0.0112 & 0.0079 & 15.70 & 15.37 & 22.61 & 9.90 & 33.84 & 18.60 \\
\midrule				
B-LSTM~\cite{wu2020pq} & 0.0131 & 0.0125 & 0.0077 & 21.77 & 28.64 & 20.78 & 6.80 & 22.56 & 14.05 \\
\midrule				
B-Complement~\cite{sung2017complementme} & 0.0241 & 0.0298 & 0.0150 & 8.78 & 2.32 & 12.67 & 9.19 & 15.57 & 26.56 \\
\midrule
~DGL-Net~\cite{huang2020generative} & 0.0091 & 0.0050 & 0.0093 & 39.00 & 49.51 & 33.33 & 23.87 & 39.96 & 41.70 \\
\midrule
~RGL-Net~\cite{narayan2022rgl} & 0.0087 & 0.0048 & \textbf{0.0072} & 49.06 & 54.16 & 37.56 & 32.26 & 42.15 & 57.34 \\
\midrule
\textbf{Ours} & \textbf{0.0054} & \textbf{0.0035} & 0.0103 & \textbf{62.80} & \textbf{61.67} & \textbf{38.68} & \textbf{48.45} & \textbf{56.18} & \textbf{62.62} \\
\bottomrule
\end{tabular}

    \label{table:sota}
\end{table*}
\begin{figure*}[t]
\centering
\includegraphics[width=0.89\textwidth]{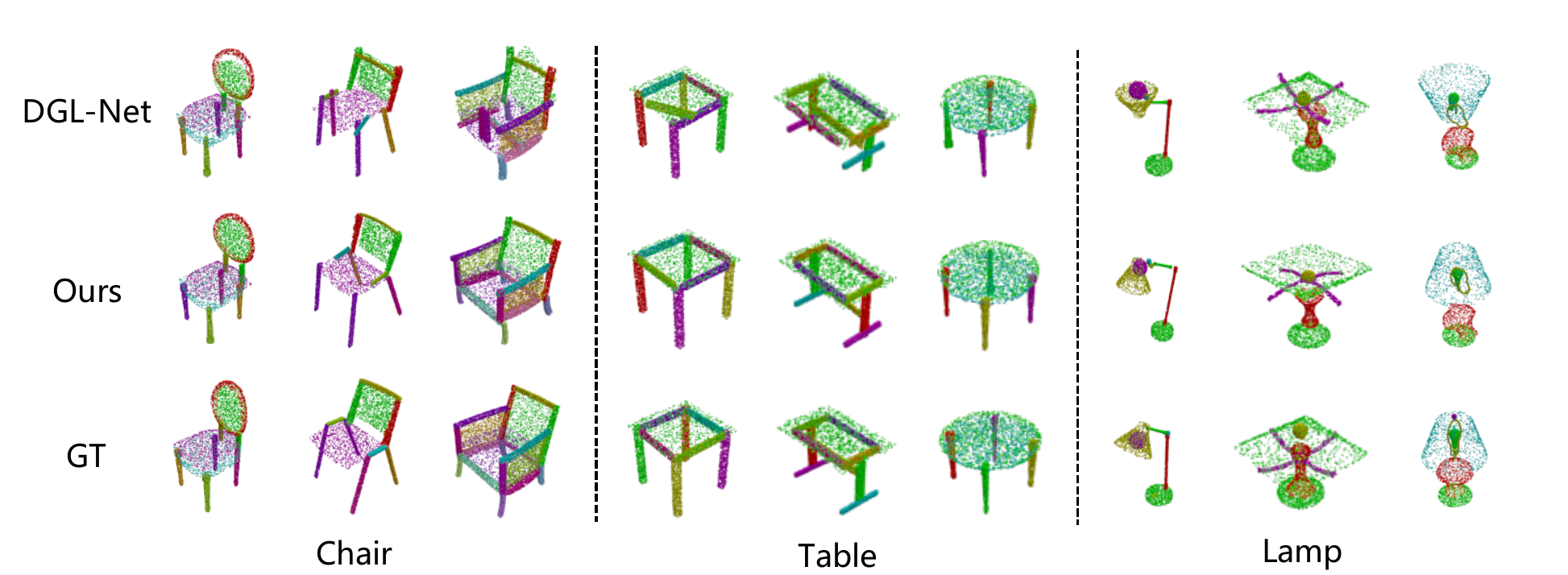}
\caption{\textbf{Qualitative results}. Our method has the ability to assemble various furniture.}
\label{fig:sota_vis}
\end{figure*}

\subsection{Main Results}

\begin{figure}[h]
\centering
\includegraphics[width=0.4\textwidth]{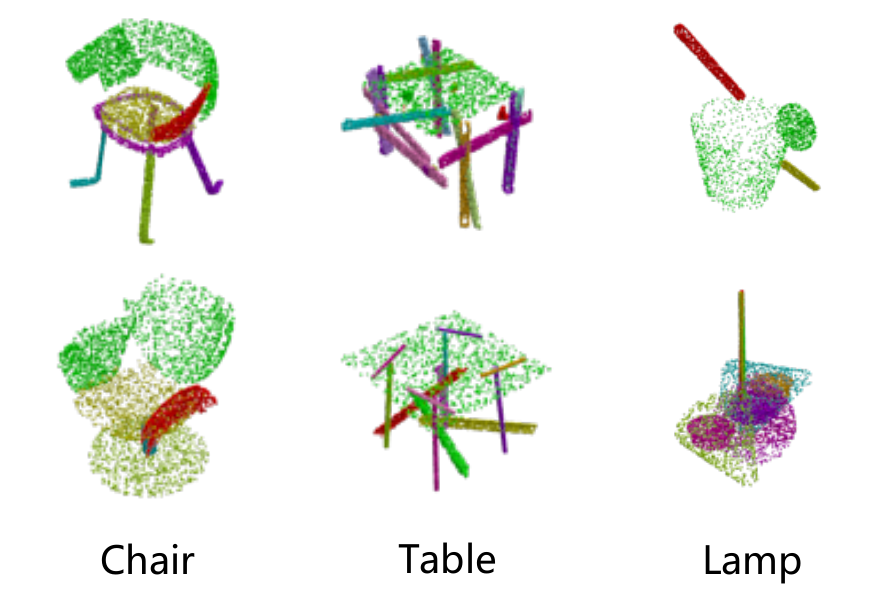}
\caption{\textbf{Some failure cases.} It is difficult to find the contact points of round-shape parts.}
\label{fig:failure}
\end{figure}

We evaluate the algorithm on PartNet~\cite{mo2019partnet} and compare our results with state-of-the-art methods in Table~\ref{table:sota}.
Our method outperforms all these approaches by a large margin, especially in the part and connectivity accuracy metrics. Specifically, we achieve 13\% PA and 16\% CA improvements on the chair category, 7\% PA and 14\% CA improvements on the table over SOTA model RGL-Net~\cite{narayan2022rgl}, individually.
Instead of time-consuming step-by-step learning part poses, we transform all parts at once.
\textit{Instance Encoding} significantly reduces the confusion between parts in parallel processing and increases the model’s capacity to distinguish indiscernible parts.
As shown in Fig.~\ref{fig:multi_poses}, we generate multiple assembled shapes for the same set of parts (with different noises) and observe that they sometimes behave differently in appearance, but all are reasonable. More visualizations are presented in Fig.~\ref{fig:sota_vis}. Our method is more likely to assemble the well-connected and structurally-stable furniture than DGL-Net~\cite{huang2020generative}.
However, the method fails to deal with round-shape components or jillion structurally-similar parts well. We detail several failing samples in Fig.~\ref{fig:failure}.

\subsection{Ablation Study}

In this subsection, a series of ablation studies are conducted to verify the effectiveness of our method. We report all results on the chair category.

\myparagraph{Instance encoding.}
As we discussed above, \textit{Instance Encoding} enables each input part uniquely while maintaining the same constraints for geometrically-equivalent parts. 
The results are shown in Table~\ref{table:ins_enc}.
The baseline model without any encoding suffers from inter-part ambiguity and fails to achieve a satisfactory result. This problem is greatly alleviated when integrating either inter-class or intra-class encoding into the network. And the performance is further improved when these two codes are employed together.

To further verify the effectiveness of \textit{Instance Encoding}, we insert it into a typical graph neural network. The model follows the basic design in~\cite{huang2020generative} and iteratively reasons the part poses and their relationships. As shown in Table~\ref{table:gnn}, The code vector boosts the performance to a large extent, which shows its generality and can be introduced into most relational-reasoning frameworks in assembly tasks.
Furthermore, we demonstrate that the transformer-based framework achieves better assembly accuracy than GNN. Unlike sparse graph prediction, \textit{Instance Encoding} is more favourable for Transformer involving dense relational reasoning.

\myparagraph{Number of encoder layers.}
We evaluate the importance of self-attention relationship learning between input parts by varying the number of encoder layers. As shown in Table~\ref{table:num_encoder}, the performance grows steadily with the increase of layers and reaches saturation at the sixth encoder layer.
The speed is reported on a single NVIDIA GTX 1080 GPU.
We choose 6 layers as the default configuration.

\myparagraph{Loss function.}
We train several models to evaluate the importance of different loss functions by turning them on and off. 
The results are listed in Table~\ref{table:loss}. 
The model performs poorly when $\mathcal{L}_{t}$ is removed, which indicates that part translation accounts for the most of assembly performance. As for the remaining losses, $\mathcal{L}_{r}$ helps learn rigid rotation for each part, and $\mathcal{L}_{s}$ ensures the holistic part assembly.

\myparagraph{Influence of noise.}
Introducing noise into the model enables us to generate multiple shape variations.
Here we monitor the impact of noise by varying the dimension of random noise and measure its quality using part accuracy metric.
Similar to~\cite{narayan2022rgl}, we use variability $V_{E}$ to quantify the variations, which is defined as the gap between best and worst performance.
And the worst assembly performance is calculated using Maximum Matching Distance, opposite to~Eq.~\eqref{eq:equ_mon}.
Mathematically, $V_{E}$ is written as:
\begin{equation}
\label{eq:equ_ve}
\footnotesize
V_{E} = \max_{j \in [E]}
\mathcal{L}(\mathcal{F}(\mathcal{P},r_{j}), \mathcal{F^{*}}(\mathcal{P})) - \min_{
j \in [E]}
\mathcal{L}(\mathcal{F}(\mathcal{P},r_{j}), \mathcal{F^{*}}(\mathcal{P}))
\end{equation}
where E is the iteration of predictions with the same input part points and we set it to 10 here. The result is presented in Fig.~\ref{fig:noise_dim}. We use the \green{solid green line} to indicate the best performance and the \blue{dashed blue line} to depict the worst matching accuracy. 
Increasing the noise dimension moderately allows our model to explore more structural varieties yet results in a decreasing infimum. We fix the noise dimension as 64 for all the furniture categories as it gives the optimal performances.

\myparagraph{Limitation of instance encoding.}
We observe that the improvement of the lamp category over previous methods is limited (\eg, 13.74\% \vs\ 1.12\% on chair and lamp PA in Table~\ref{table:sota}).
\textit{Instance Encoding} is designed to tackle the problem of inter-part confusion. However, as shown in Fig.~\ref{fig:sota_vis}, most part geometries in a lamp are easily distinguished from each other, thus weakening the effect of \textit{Instance Encoding}.

\begin{table}[t!]	
	\centering
	\caption{The effect of instance encoding.}
    \setlength{\tabcolsep}{4.8mm}
\small 
\centering
\begin{tabular}{c|c|c|c}
encoding? & SCD $\downarrow$ & PA $\uparrow$ & CA $\uparrow$ \\
\Xhline{2\arrayrulewidth}
\rule{0pt}{1.1\normalbaselineskip}
- & 0.0068 & 48.40 & 36.04 \\
inter-only & 0.0055 & 59.93 & 42.66 \\
intra-only & 0.0059 & 60.78 & 45.92 \\
ins-enc & \textbf{0.0054} & \textbf{62.80} & \textbf{48.45} \\
\end{tabular}
    \label{table:ins_enc}
\end{table}

\begin{table}[t!]	
	\centering
	\caption{Different architecture with instance encoding.}
    \setlength{\tabcolsep}{3mm}
\small 
\centering
\begin{tabular}{c|c|c|c|c}
arch & w/ins-enc? & SCD $\downarrow$ & PA $\uparrow$ & CA $\uparrow$ \\
\Xhline{2\arrayrulewidth}
\rule{0pt}{1.1\normalbaselineskip}
\multirow{2}*{GNN} & ~ & 0.0068 & 46.44 & 33.77 \\
~ & \cmark & 0.0063 & 58.14 & 43.19 \\
\hline
\rule{0pt}{1.0\normalbaselineskip}
\multirow{2}*{Transformer} & ~ & 0.0068 & 48.40 & 36.04 \\
~ & \cmark & \textbf{0.0054} & \textbf{62.80} & \textbf{48.45} \\
\end{tabular}
    \label{table:gnn}
\end{table}

\begin{table}[t!]	
	\centering
	\caption{Ablation study of encoder depth.}
    \setlength{\tabcolsep}{4.2mm}
\small 
\centering
\begin{tabular}{c|c|c|c|c}
layers & SCD $\downarrow$ & PA $\uparrow$ & CA $\uparrow$ & FPS \\
\Xhline{2\arrayrulewidth}
\rule{0pt}{1.1\normalbaselineskip}
2~ &	0.0065 & 56.10 & 34.37 & \textbf{30.5} \\
3 &	0.0061 & 58.66 & 37.20 & 27.1 \\
4 &	0.0057 & 61.29 & 41.08 & 24.6 \\
6 &	\textbf{0.0054} & \textbf{62.80} & \textbf{48.45} & 21.2 \\
8 &	0.0055 & 62.30 & 46.75 & 19.1 \\
\end{tabular}
    \label{table:num_encoder}
\end{table}


\begin{table}[t!]	
	\centering
	\setlength{\abovecaptionskip}{0cm}
	\caption{The influence of loss components.}
    \setlength{\tabcolsep}{3.6mm}
\small 
\centering
\begin{tabular}{c c c|c|c|c}
$\mathcal{L}_{t}$ & $\mathcal{L}_{r}$ & $\mathcal{L}_{s}$ & SCD $\downarrow$ & PA $\uparrow$ & CA $\uparrow$ \\
\Xhline{2\arrayrulewidth}
\rule{0pt}{1.1\normalbaselineskip}
 & \checkmark  & \checkmark &  0.0070 & 29.04 & 29.44 \\
\checkmark & & \checkmark  & 0.0056 & 59.51 & 43.07 \\
\checkmark  & \checkmark &  & 0.0058 & 61.07 & 45.39 \\
\checkmark  & \checkmark & \checkmark & \textbf{0.0054} & \textbf{62.80} & \textbf{48.45} \\
\end{tabular}
    \vspace{-0.12cm}
    \label{table:loss}
\end{table}

\begin{figure}[!t]
\centering
\setlength{\abovecaptionskip}{-0.05cm}
\includegraphics[width=0.48\textwidth]{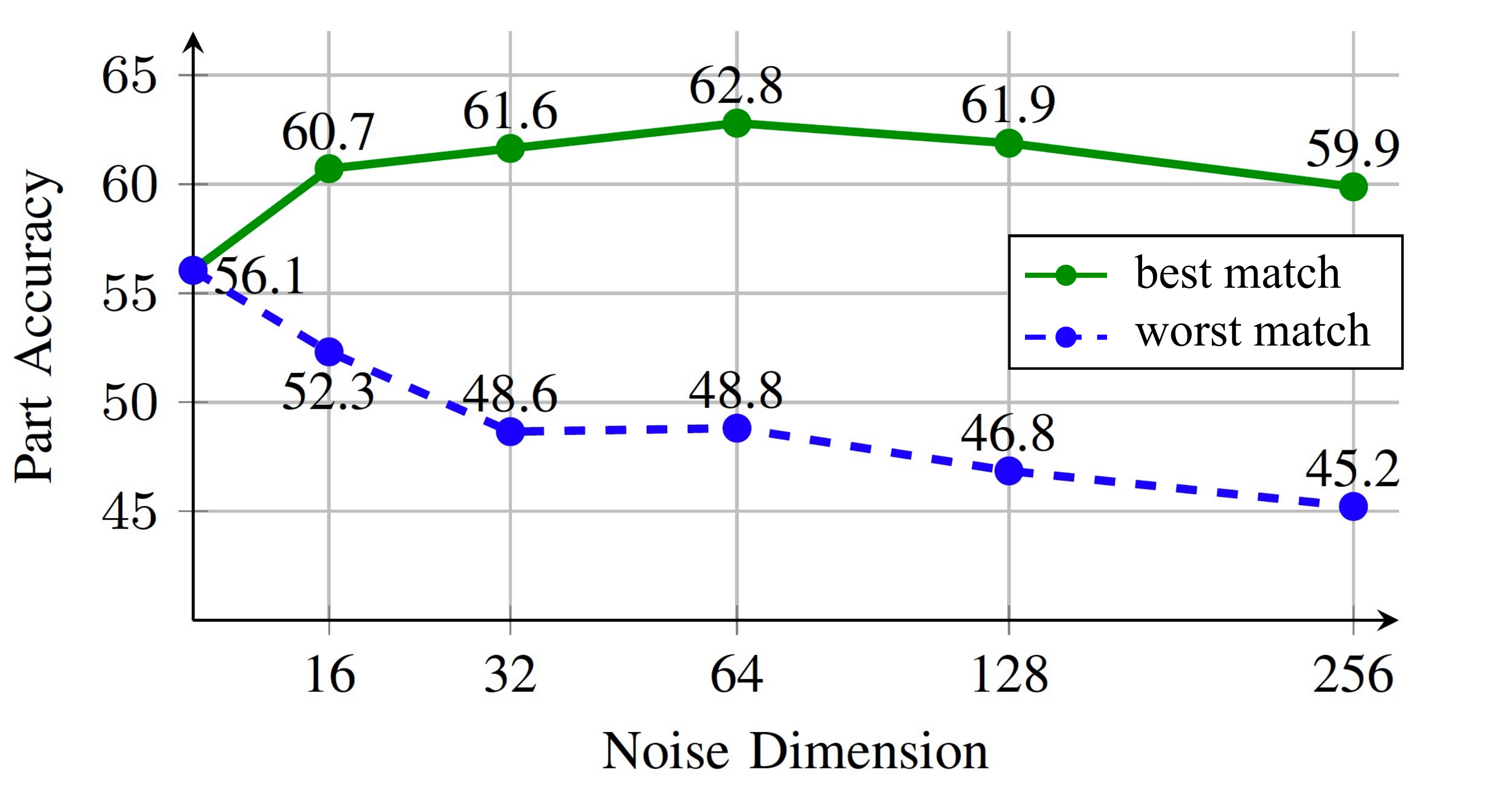}
\caption{\textbf{Performance of our network on the varying dimension of random noise}. We still achieve the best performance even without MoN inference (56.1\% at zero noise \vs\ 49.1\% in RGL-Net~\cite{narayan2022rgl}).}    
\vspace{-0.12cm}
\label{fig:noise_dim}
\end{figure}

\subsection{In-process Part Assembly Task}
The task requires 1) understanding the global posture of in-process furniture, 2) reasoning the relationship between furniture and scattered parts, and 3) placing the candidate parts in their appropriate locations. And this function will significantly generalize the applicability in reality, such as furniture maintenance. Given a broken chair, instead of dismantling a series of parts and then re-assemble the furniture as part assembly task, we can replace the damaged part with a new one directly. 
Our framework has the potential for this task using an encoder-decoder paradigm. The unfinished furniture is fed to encoder to generate feature memory and the remaining parts query the information from the memory through the decoder. 
Compared to progressive assembly~\cite{narayan2022rgl} involving the computational complexity of $\mathcal{O}(n^2)$, the complexity in our memory-query paradigm decreases to $\mathcal{O}(n)$, which dramatically improves assembly efficiency.

We employ \textit{PartDrop} scheme to enhance its ability to extract features from incomplete shapes. 
The input part is randomly dropped with a certain probability during training.
To evaluate the model's capacity to reason about in-process inventory, we sequentially remove one part from the input parts to generate the incomplete shape during the inference phase. 
We calculate the average of all one-part-removed candidates in a complete shape as the evaluation result.
Moreover, the complete 3D part clouds in the regular assembly task is termed as regular shape.
The overall results are presented in Table~\ref{table:partdrop}.
We set the \texttt{drop-prob} to 0.2 as it works well on both incomplete and regular shapes.
Interestingly, we observe it is also beneficial to regular part assembly, particularly part connection accuracy ($\sim$8\% improvement). 
It is essentially a kind of data augmentation and generates more complex data.
This operation helps our model to better understand fine-grained part connections.
We evaluate the quality of in-process parts through Eq.~\eqref{eq:equ_wip} and achieve 62.36/52.95 in PA and CA, individually.

\begin{table}[t!]	
	\centering
	\caption{The importance of Part Drop in in-process assembly.}
    \setlength{\tabcolsep}{1.6mm}
\small 
\centering
\begin{tabular}{c|ccc|ccc}
\toprule
\multirow{2}{*}{drop-prob} & \multicolumn{3}{c}{Incomplete Shape} & \multicolumn{3}{c}{Regular Shape} \\
 & SCD $\downarrow$ & PA $\uparrow$ & CA $\uparrow$ 
& SCD $\downarrow$ & PA $\uparrow$ & CA $\uparrow$ \\
\Xhline{2\arrayrulewidth}
\rule{0pt}{1.1\normalbaselineskip}
0~ & 0.0093 & 50.24 & 37.44 & 0.0059 & 61.33 & 40.38 \\ 
0.1 & 0.0075 & 54.23 & 42.44 & 0.0055 & 61.76 & 45.65 \\
0.2 & 0.0071 & 55.52 & \textbf{45.46} & 0.0054 & \textbf{62.80} & \textbf{48.45} \\
0.5 & \textbf{0.0068} & \textbf{55.70} & 44.78 & \textbf{0.0053} & 61.73 & 48.04 \\
\end{tabular}

    \vspace{-0.2cm}
    \label{table:partdrop}
\end{table}

\section{CONCLUSIONS}

In this work, we present an instance-aware relational-reasoning framework for part assembly.
We establish new state-of-the-art results in the regular assembly task and further introduce a practical problem called in-process assembly. Given the simplicity, efficiency, and firm performance, we hope that our method can serve as a cornerstone for many automated assembly tasks. Considering data efficiency and interactive safety, present works mainly conduct assembly in simulation and are still inapplicable in reality. In future work, we are intended to bridge the gap between simulation and reality, and transfer the gained assembly skill to accomplish more complicated tasks in the real world.




\bibliographystyle{IEEEtran}
\bibliography{IEEEabrv, reference}

\end{document}